%% file: main.tex
\pdfoutput = 1
\documentclass[onecolumn]{article}
\usepackage{url}
\usepackage{hyperref}
\usepackage{multirow}
\usepackage[sort,numbers,sort&compress]{natbib}
\usepackage{float}
\usepackage{placeins}
\usepackage{subcaption}
\usepackage{graphicx}
\usepackage{pifont}
\usepackage{multirow}
\usepackage{placeins}
\usepackage{subcaption}
\usepackage{graphicx}
\newsavebox{\foobox}
\usepackage{float}
\usepackage{afterpage}
\usepackage{booktabs}
\usepackage{colortbl}
\usepackage{xcolor}
\usepackage{amsmath, amsthm, amssymb, bbm, bm}
\usepackage{courier}
\newcommand{\fakeslant}[2][0.25em]{\makebox[2.5pt][l]{#2}\kern-#1#2}
\newcommand{\boldcourier}[1]{\textbf{\texttt{\fakeslant{#1}}}}

\newcommand\maxboxacc{\texttt{MaxBoxAcc}\xspace}
\newcommand\newmaxboxacc{\texttt{MaxBoxAccV2}\xspace}

\newcommand\locbold{\boldcourier{\fontfamily{lmtt}\fontseries{b}\selectfont\xspace loc.}\xspace}

\newcommand\backbonebold{\boldcourier{\fontfamily{lmtt}\fontseries{b}\selectfont Backbone}\xspace}
\newcommand\newmaxboxaccbold{\boldcourier{\fontfamily{lmtt}\fontseries{b}\selectfont MaxBoxAccV2}\xspace}

\newcommand\pxap{\texttt{PxAP}\xspace}
\newcommand\pxapbold{\boldcourier{\fontfamily{lmtt}\fontseries{b}\selectfont PxAP}\xspace}
\newcommand\cls{\texttt{class}\xspace}
\newcommand\clscapt{\texttt{Class}\xspace}
\newcommand\topone{\texttt{top-1}\xspace}
\newcommand\topfive{\texttt{top-5}\xspace}
\newcommand\toponebold{\boldcourier{\fontfamily{lmtt}\fontseries{b}\selectfont Top-1}\xspace}
\newcommand\topfivebold{\boldcourier{\fontfamily{lmtt}\fontseries{b}\selectfont Top-5}\xspace}

\newcommand\lpebold{\boldcourier{\fontfamily{lmtt}\fontseries{b}\selectfont LPE}\xspace}

\newcommand\lmebold{\boldcourier{\fontfamily{lmtt}\fontseries{b}\selectfont LME}\xspace}
\newcommand\mie{\texttt{MIns-Error}\xspace}
\newcommand\miebold{\boldcourier{\fontfamily{lmtt}\fontseries{b}\selectfont MIns-Error}\xspace}

\newcommand\ioa{\texttt{IoA}\xspace}
\newcommand\iop{\texttt{IoP}\xspace}
\newcommand\iog{\texttt{IoG}\xspace}
\newcommand\iou{\texttt{IoU}\xspace}

\definecolor{darkergreen}{RGB}{21, 152, 56}
\definecolor{red2}{RGB}{252, 54, 65}
\definecolor{Gray}{gray}{0.85}
\newcolumntype{g}{>{\columncolor{Gray}}c}

\definecolor{greenbg}{RGB}{168,199,126}
\definecolor{bluebg}{RGB}{125,200,255}

\newcommand*{\ie}{\emph{i.e.,}\@\xspace}
\newcommand{\abs}[1]{\ensuremath \left| #1 \right|}
\newcommand{\reals}{\mathbb{R}}

\usepackage[boxruled, vlined, linesnumbered]{algorithm2e}
\SetAlFnt{\small}
\SetAlCapFnt{\small}
\SetAlCapNameFnt{\small}
\usepackage{algorithmic}
\algsetup{linenosize=\tiny}

\definecolor{mybluelight}{rgb}{0.9, 0.9, 1.}

\usepackage{style}

\makeatletter
\@namedef{ver@everyshi.sty}{}
\newcommand{\removelatexerror}{\let\@latex@error\@gobble}

\title{DiPS: \textbf{Di}scriminative \textbf{P}seudo-Label \textbf{S}ampling with Self-Supervised Transformers for Weakly Supervised Object Localization}

\renewcommand\footnotemark{}

\author{Shakeeb~Murtaza$^{1}$,
  ~Soufiane~Belharbi$^{1}$,
  ~Marco~Pedersoli$^{1}$,
  ~Aydin~Sarraf$^{2}$, and
  ~Eric~Granger$^{1}$\\
 $^1$ LIVIA, Dept. of Systems Engineering, ETS Montreal, Canada \\
$^2$  Ericsson, Global AI Accelerator, Montreal, Canada\\
{\tt\footnotesize \textcolor{black}{shakeeb.murtaza.1@ens.etsmtl.ca} }
}

\newcommand{\ignore}[1]{}

 \usepackage[shortlabels]{enumitem}

\begin{document}
\maketitle\thispagestyle{fancy}

\maketitle
\lhead{\color{gray} \small }
\rhead{\color{gray} \small Murtaza et al. \;  [Image and Vision Computing 2023]}

\begin{abstract}
Self-supervised vision transformers (SSTs) have shown great potential to yield rich localization maps that highlight different objects in an image. However, these maps remain class-agnostic since the model is unsupervised. They often tend to decompose the image into multiple maps containing different objects while being unable to distinguish the object of interest from background noise objects. In this paper, \textbf{Di}scriminative \textbf{P}seudo-label \textbf{S}ampling (DiPS) is introduced to leverage these class-agnostic maps for weakly-supervised object localization (WSOL), where only image-class labels are available. Given multiple attention maps, DiPS relies on a pre-trained classifier to identify the most discriminative regions of each attention map. This ensures that the selected ROIs cover the correct image object while discarding the background ones, and, as such, provides a rich pool of diverse and discriminative proposals to cover different parts of the object. Subsequently, these proposals are used as pseudo-labels to train our new transformer-based WSOL model designed to perform classification and localization tasks.  Unlike standard WSOL methods, DiPS optimizes performance in both tasks by using a transformer encoder and a dedicated output head for each task, each trained using dedicated loss functions. To avoid overfitting a single proposal and promote better object coverage, a single proposal is randomly selected among the top ones for a training image at each training step.  Experimental results\footnote{Our code is available: \href{https://github.com/shakeebmurtaza/dips}{https://github.com/shakeebmurtaza/dips}} on the challenging CUB, ILSVRC, OpenImages, and TelDrone datasets indicate that our architecture, in combination with our transformer-based proposals, can yield better localization performance than state-of-the-art methods.
\end{abstract}

\section{Introduction}
\label{sec:intro}

The recent success of deep learning models (DL) in different visual recognition tasks, such as image classification~\citep{he2016deep}, object localization~\citep{choe2020evaluating}, detection~\citep{RedmonDGF16CVPR}, and segmentation~\citep{ChenPKMY18} requires big models, and most importantly, large, annotated datasets. The high cost of dense localization and of segmentation supervision makes it difficult to train these models and annotate large-scale datasets. Weakly supervised object localization (WSOL) has recently emerged as a surrogate training strategy to alleviate the need for bounding box supervision~\citep{zhou2016learning}. This allows a cost-effective and fast collection of large datasets. Using only a global image label, \ie image class, a DL model can be trained to classify an image and localize objects of interest.

\begin{figure}[hb!]\captionsetup[subfigure]{font={sf},labelfont={bf,sf}}
\centering
\begin{subfigure}[b]{0.6\textwidth}
\begin{center}
   \includegraphics[width=0.8\linewidth]{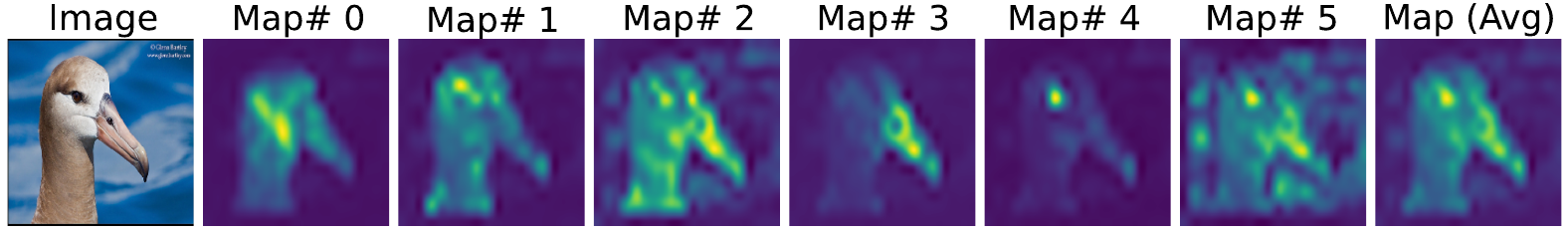}
   \caption{\clscapt tokens from a trained self-supervised transformer (SST) \citep{caron2021emerging}.}
   \label{fig:dino_last_head_subfig}
\end{center}
\end{subfigure}
\centering
\vspace{3mm}

\begin{subfigure}[b]{0.6\textwidth}
\begin{center}
   \includegraphics[width=0.8\linewidth]{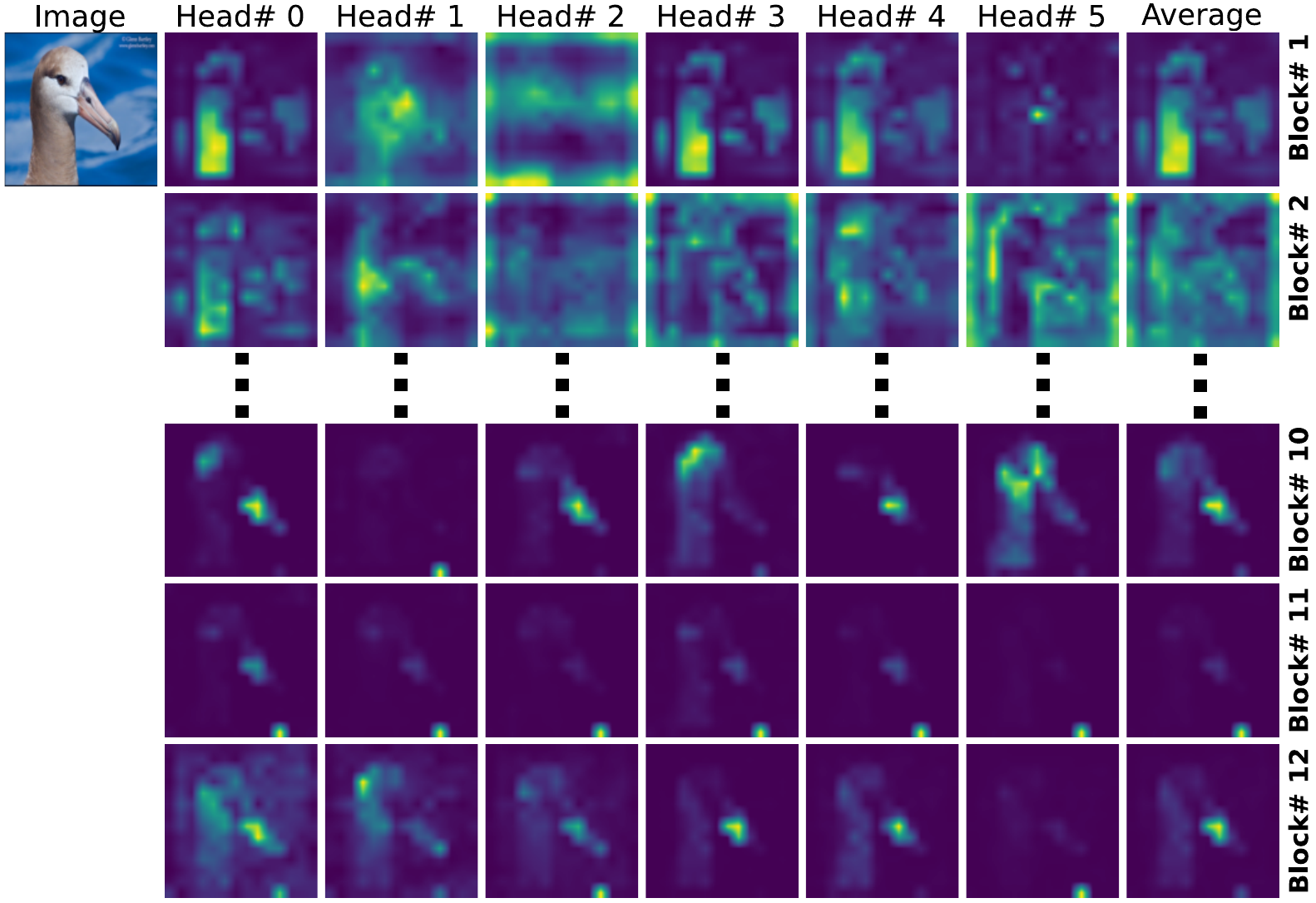}
   \caption{\clscapt tokens from TS-CAM method~\citep{gao2021ts} trained using image-label.}
   \label{fig:tscam_maps_subfig} 
   \end{center}
\end{subfigure}
\vspace{3mm}

\begin{subfigure}[b]{0.6\textwidth}
\begin{center}
   \includegraphics[width=0.8\linewidth]{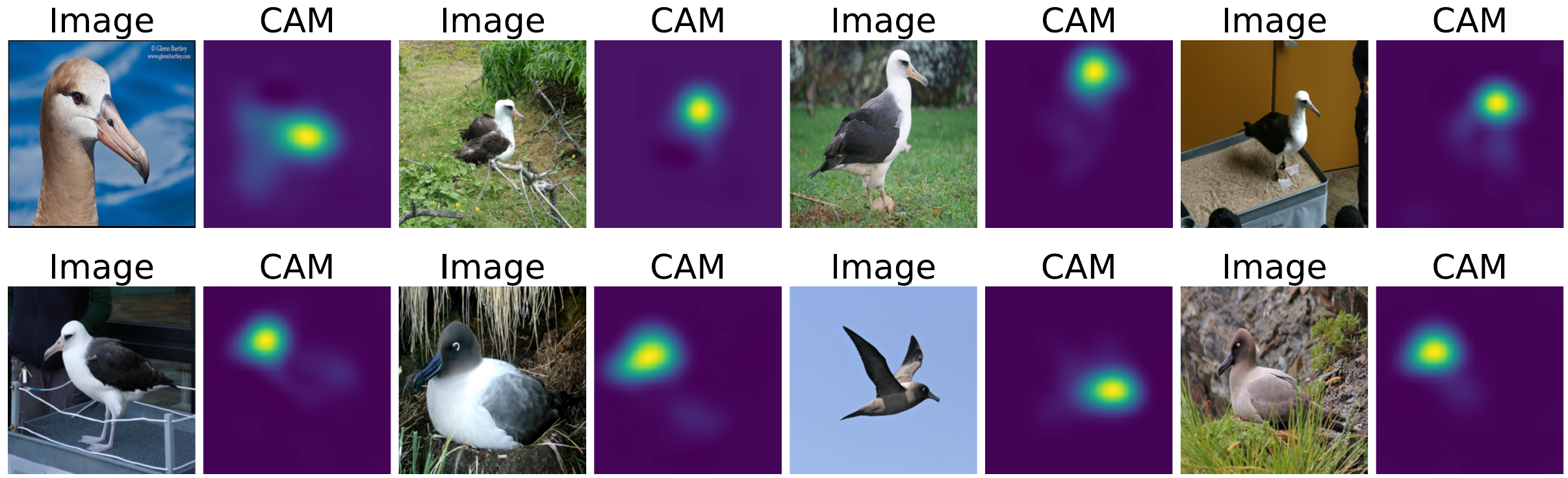}
   \caption{Class-activation Maps (CAMs)~\citep{zhou2016learning}.}
   \label{fig:camfromcnn}
\end{center}
\end{subfigure}
\caption[]{
Comparison of activation maps of three different models: 
\textbf{(a):} The last layer of the SST attends to all parts of foreground regions. 
\textbf{(b):} Visualizations of \cls tokens collected from different attention layers of TS-CAM \citep{gao2021ts}. Earlier attention layers attend to background regions instead of focusing exclusively on foreground regions. Fusing all these maps, as in TS-CAM, introduces noisy localization.
\textbf{(c):} CAMs collected from CNN-based model~\citep{oquab2015object} show that they are very local and only focus on a small discriminative area since correctly classifying the image is deemed to be enough. Note that large parts of the object are missing, compared to SST maps. Naturally, this positions SST maps as a better choice for sampling good-quality pseudo-labels.
}
\label{fig:dino_last_head}
\end{figure}

Class activation Mapping (CAM) methods are Convolution Neural Network (CNN)-based approaches that have been dominating the WSOL field~\citep{belharbi2023colocam,belharbi2022fcam,choe2019attention,lee2019ficklenet,rahimi2020pairwise,singh2017hide,wei2021shallow,wei2017object,xue2019danet,yang2020combinational,yun2019cutmix,zhang2020rethinking,zhang2018self}. They require only image-class supervision, and leverage spatial information in a CNN. Guided only by a discriminative loss, CAM methods can yield a spatial map per class to localize an object of interest while classifying the image. However, this allows the emergence of only the most discriminative part of an object~\citep{choe2020evaluating,rony2023deep}. Often, it is limited to small repetitive patterns across samples of a class. Such limited coverage leads to poor localization since large parts of an object are missing (Fig.\ref{fig:camfromcnn}).
Different techniques have been proposed to improve CAM-based localization~\cite{rony2023deep}, including different spatial poolings, data augmentation, feature enhancement, and feedback methods. A recent prominent line of research aims to leverage pseudo-labels to fine-tune models~\cite{ZhangCW20rethink,wei2021shallowspol,negevsbelharbi2022,belharbi2022fcam,tcamsbelharbi2023,belharbi2023colocam}. Various works have shown that classification and localization tasks are antagonistic in a WSOL setup~\cite{ZhangCW20rethink,choe2020evaluating,rony2023deep}, where localization converges very early on, while classification converges late in the training. Typically, a classifier is trained until convergence and used to generate pseudo-labels. Then, it is frozen and equipped with a localization branch to be fine-tuned for localization. This allows to build a single model that yields the best performance over both tasks. Despite the success of pseudo-labeling methods, the performance obtained is strongly tied to the emerging localization, \ie ROIs, in CAMs, which tend to be \emph{local}, and limited to a small part of an object. This makes the pseudo-labels less efficient.

Recently, self-supervised transformers (SSTs)~\citep{caron2021emerging} have come to constitute a strong competitor to CNN-based models. Without any supervision, they can produce good attention maps for different objects in an image (Fig.\ref{fig:dino_last_head_subfig}).
With their long-range dependency and self-supervised training, SST models can decompose different objects into multiple maps. However, they are unable to discriminate between different objects linked to a particular class. Most importantly, though, they can accumulate the \emph{localization} information of different objects into multiple attention maps (tokens). These maps highlight all objects in a scene as they are trained without any class information. Each token focuses on different objects that are semantically distinct from one another. Although this provides a rich source for object localization, these attention maps are not associated with any particular class, making them less useful for WSOL tasks.

\begin{figure*}[!t]
\begin{center}
 \includegraphics[width=1\linewidth,trim=0 0 0 0, clip]{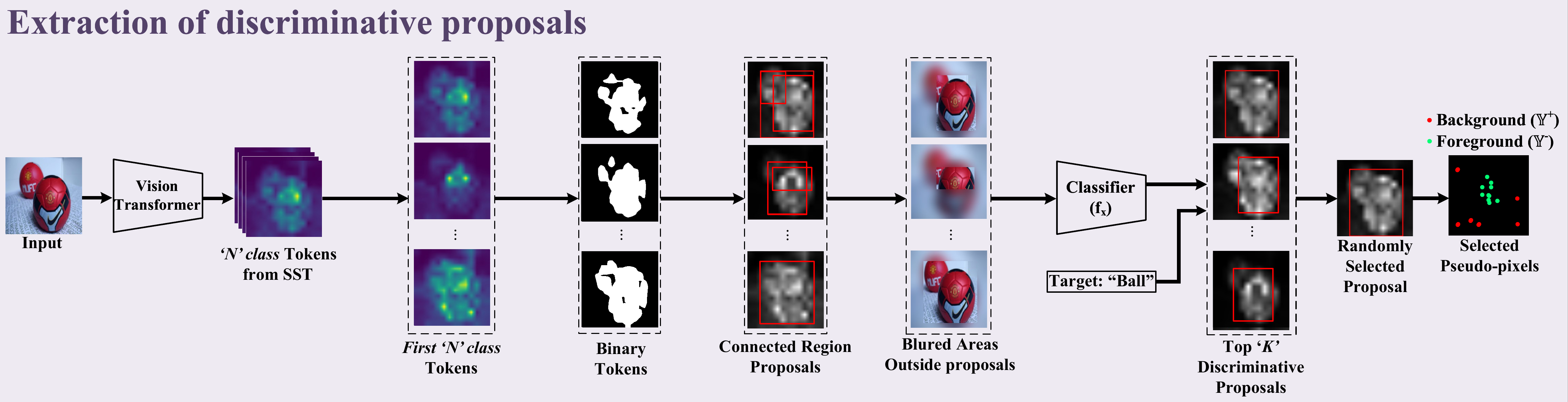}
\end{center}
\caption{\textit{Our approach for generating \textbf{discriminative} proposals from \cls tokens of SSTs.} First,  the localization map are automatically binarized~\citep{otsu1979threshold} to identify regions of interest that may contain a specific object. From each binarized map, all connected areas are extracted along with their tight bounding boxes. This creates a pool of bounding boxes from different attention maps. Each bounding box is scored using an external classifier ${f_x}$. We use the posterior probability of the true class $y$ of the image to measure the likelihood of the bounding box containing the true object. Outside of the box is suppressed by blurring operations. The top-${K}$ boxes are selected.
At each training step, we randomly select a proposal box. Inside the box, we randomly select foreground pixels. While background pixels are selected from outside the box.
A detailed flow diagram is displayed in Fig.\ref{fig:our_method}
\label{fig:pseudo_label}.
}
\end{figure*}

Various works have attempted to leverage transformers for WSOL~\citep{bai2022weakly, chen2022lctr, gao2021ts, gupta2022vitol, su2022re} by using the image class as supervision. The TS-CAM method~\citep{gao2021ts} is one of the prominent transformer-based WSOL approaches. It performs a simple fusion of all attention maps across all layers to build a single class-agnostic map. This map is then aggregated with a CAM for final localization. Such cross-layer aggregation of maps introduces localization noise (Fig.\ref{fig:tscam_maps_subfig}), which hampers performance.  In~\citep{bai2022weakly,gupta2022vitol}, the authors propose to suppress background noise and calibrate foreground activations. Despite the success of these methods, they still need complex changes to deal with background noise. Building class-aware attention maps in transformer-based models is still an ongoing challenge.


In this work, we aim to design a method to exploit the rich localization information built into self-supervised transformers. Unlike TS-CAM~\citep{gao2021ts}, which fuses attention maps, we propose to use a pre-trained classifier to identify the most \emph{discriminative ROIs}, with respect to the true image class, across different attention maps (Fig.\ref{fig:pseudo_label}). This allows us to gather more reliable and diverse proposals that better cover the object class in the image. Furthermore, it ensures that background noise objects, which typically emerge in such maps, are discarded.  
We adopt a fine-tuning approach for WSOL to build a model that achieves the best results in both the classification and localization tasks. More specifically, we leverage self-supervised transformers by using them as a backbone and employ the aforementioned proposals to train them. The localized discriminative proposals from an SST are used to sample pixel-wise pseudo-labels, which are then used to train our model for localization. These pseudo-labels, as elaborated in our preliminary study \cite{Murtaza_2023_WACV}, are important in the training of the localization model. Moreover, our final model is composed of a transformer-based encoder equipped with two output heads, namely, a classification head and a localization head. Both are trained separately for better performance.


\noindent \textbf{Our main contributions are summarized as follows:}\label{summ:contributions}

\noindent \textbf{(1)} A novel method called Discriminative Pseudo-Label Sampling (DiPS) is introduced to leverage the rich localization information contained in a set of \emph{class-agnostic} attention maps of SSTs (see Fig.\ref{fig:pseudo_label}). Using a pre-trained classifier, DiPS collects the most discriminative ROIs with respect to the image class, while discarding the background noise objects. Given the multiple attention maps produced by transformers and different ROIs, our method provides a rich pool of diverse and discriminative proposals to cover different parts of the object. DiPS allows for the production of reliable pseudo-labels with better object coverage during localization. This differs from standard CAMs, which are constrained to highlighting smaller object regions by minimizing the mutual information between class instances.\\
\textbf{(2)} Our new DiPS method is trained to simultaneously provide a high level of classification and localization accuracy. It is composed of a transformer-based encoder, a classification head, and a localization head. The encoder is pre-trained under an SST model, which is frozen. The classification head is trained to yield the best classification accuracy and then is frozen as well. Moreover, the localization head is trained using the discriminative pseudo-labels harvested from SST. To avoid overfitting a single proposal and promote better object coverage, DiPS randomly selects a single proposal among the top ones for a training image at each training step. Additionally, for a better object-boundary delimitation, it uses a CRF loss~\citep{tang2018regularized}.  Moreover, \textit{DiPS} only requires a single forward pass through the encoder and respective heads to concurrently accomplish classification and localization tasks. \\
\textbf{(3)} Our extensive results compare DiPS with state-of-the-art WSOL methods on four datasets -- three public common benchmarks for the WSOL task (ILSVRC, CUB-200-2011, OpenImages), and a fourth proprietary dataset, TelDrone, designed for cell tower inspection using drones. Our DiPS method outperforms recent WSOL methods and achieves a new state-of-the-art localization performance. Further analysis and ablations are provided for our method, along with our public code.

\section{Related Work}\label{sec:relatedwork}
This section provides a brief summary of WSOL methods built on top of CNNs and vision transformers (ViTs).

\noindent\textbf{Convolutional Neural Network:} State-of-the-art WSOL methods are designed to collect the localization map from the last convolution layer of CNNs~\citep{zhou2016learning}. These maps can be generated by aggregating the activation maps from the penultimate layer based on the contribution of each map toward the final prediction. Different methods have been proposed to improve the map extraction mechanism for WSOL~\citep{chattopadhay2018grad,fu2020axiom, ramaswamy2020ablation,selvaraju2017grad}. These methods always focus on discriminative regions as they are optimized using class-level labels. This allows the network to find the common object parts between instances of the same class. To address this limitation, different methods have been proposed to expand the receptive field beyond discriminative regions. Among them, the most common one used for enlarging the activation map beyond discriminative regions proceeds by removing them either by an adversarial perturbation~\citep{singh2017hide,yun2019cutmix} or by employing an adversarial loss~\citep{choe2019attention,zhang2018adversarial}. On the other hand, instead of carrying out weighted averaging of all the activation maps of a specific layer, the proposed fusion-based methods combine the activation maps according to their importance as determined by the classifier’s score~\citep{naidu2020cam, wang2020ss, wang2020score}.

Instead of using post-hoc techniques for collecting the localization maps from pre-trained networks, different architectures have been designed for WSOL that are able to generate localization maps directly instead of fusing activation maps~\citep{lee2019ficklenet, rahimi2020pairwise, xue2019danet, yang2020combinational, zhang2020rethinking}. For instance, \citep{wei2018revisiting} replaces the penultimate convolution layer with a layer having multiple parallel convolution filters with different dilation rates. To generate the localization maps, standard CAMs are added to the average of different CAMs generated from dilated convolution layers. The optimization of this network is extremely difficult as the classification loss must be minimized by using a separate classification for each convolution layer in the penultimate layers. For instance, \citep{yang2020combinational} proposes combining CAMs of different classes from highest to lowest, based on class probability scores. This method is also cost-intensive as CAMs must be averaged from different classes. To overcome this limitation, \citep{wu2021background, zhu2021background} proposes a method to suppress the background regions to help the network identify foreground regions with high confidence.

In \citep{meng2021foreground}, the author introduces object-aware and part-aware attention modules to jointly optimize localization and classification accuracy. This model first obtains the whole object and then decomposes it into parts for classification. It is capable of producing efficient localization maps while retaining the same classification performance. Additionally, \citep{xie2021online} added a sub-network between different layers of the network to regularize its internal features. The sub-network is an encoder-decoder with a classification head serving to preserve the details of objects at different levels in the network. In \citep{zhang2018self}, the author proposes self-produced guidance (SPG) for expanding the foreground map beyond the discriminative regions. SPG first collects seeds from the attention map, which are then expanded to areas having high activation values to produce the foreground map. Inter-image communication (I$^2$C) \citep{zhang2020inter} was introduced to expand the activation maps to cover the whole object. Nonetheless, this approach adversely affects classification performance. To address this issue, \citep{pan2021unveiling} proposed a model with two modules, namely, an activation-restricted module (RAM) to discover the parts of the objects by using a classification network and a self-correlation map-generating (SCG) module for producing the final map. Shallow feature-aware pseudo-supervised object localization (SPOL) produces pseudo-labels with high confidence for background and foreground regions~\citep{wei2021shallow}. However, these pseudo-labels remain fixed for different iterations, which could generate noisy maps. Hence,~\citep{belharbi2022fcam, Murtaza_2023_WACV} propose using effective pseudo-labels that will be sampled probabilistically at each step, allowing the network to discover related parts of the image to improve the localization accuracy.

The aforementioned methods rely on the activation maps of CNNs trained using classification loss, which limits their receptive field. Inductive bias forces the CNNs to decompose the object into local semantic parts \citep{bau2017network, zeiler2014visualizing}, preventing them from forming global relationships between object parts presented in different receptive fields. This hinders the ability of CNNs (trained using a classification loss) to detect all the object parts, which thus leads to blobby maps. Furthermore, several approaches have been proposed to extend the localization map beyond merely discriminative regions. These methods harness pixel similarities to allow the network to identify various parts of an object~\citep{wang2018weakly, wang2020self, zhang2020rethinking, zhang2020inter}. Concurrently, \citep{ki2020insample} proposed to leverage long-range dependencies in CNNs to capture spatial similarities, thereby enhancing object coverage.

\noindent\textbf{Vision Transformer (ViT) Methods:} In contrast to CNNs, transformer networks have the ability to capture long-range dependencies due to their intrinsic properties, which help generate efficient localization maps. These models rely heavily on attention mechanisms by computing the dot product between key-value pairs at different levels. Recently, the transformers were effectively employed for computer vision tasks, achieving great success. Specifically, various transformer-based methods for WSOL have been proposed~\citep{chen2022lctr, li2022caft, gupta2022vitol, bai2022weakly, su2022re, meng2022adversarial}. TS-CAM is another emerging work for WSOL using a transformer~\citep{gao2021ts}. It employs a classification head on top of the \cls tokens to train a transformer using class labels and then extracts a semantic-aware map from it. The map is then multiplied by the average of all \cls tokens from different layers. This accumulation introduces background noise, which hinders the performance of this method. Moreover, TS-CAM focuses solely on capturing long-range dependencies while ignoring the inductive locality bias, leading to unreliable maps. To deal with this, \citep{bai2022weakly} proposes a calibration mechanism for calibrating the network to produce relatively smooth activation values for different objects. This mechanism limits the receptive field of the transformer, producing more stable activation values across different parts of an object of interest. It also introduces the Spatial Calibration Module (SCM) to align the object boundaries with the edges of the localization map. Similarly, \citep{chen2022lctr} introduces a local continuity transformer (LCTR) for fusing local and global features to improve the perception quality of activation maps for an object of interest. These features are fused by incorporating two modules into the network, namely, the relational patch-attention module (RPAM) and the cue-digging module (CDM). RPAM and CDM help in retaining the global features and highlighting the less discriminative object parts, respectively. Furthermore, to reduce the background noise in the localization map generated by transformers, \citep{gupta2022vitol} presents a novel dropout mechanism within a transformer block to limit the receptive field of the transformers. This method utilizes the attention roll-out method \citep{abnar2020quantifying} to produce the attention maps for a particular object, significantly mitigating the background noise. 
Similarly, \citep{su2022re} proposes a token refinement transformer (TRT) for WSOL to produce high-confidence localization maps. Here, the TRT employs a token priority scoring module (TPSM) to capture the precise object semantics by suppressing the background regions.
 
The aforementioned methods require an optimal threshold to draw bounding boxes around the object of interest, making them sensitive to threshold values. Also, most baseline methods do not minimize the loss over generated maps, which results in blobby and unreliable localization maps. Although some methods use pseudo-labels to train the localization module, they fix pseudo-labels and force the underlying model to generate maps close to the pseudo-label~\citep{wei2021shallow}. Similarly,~\citep{belharbi2022fcam} employed a probabilistic sampling for building pseudo-labels from standard CAMs, restricting their performance because CAMs always highlight mutual information common to different instances of a particular class. Moreover, transformer-based methods accumulate attention maps from different layers and fuse them with activation maps, introducing a background noise in the final map~\citep{gao2021ts}. Comparatively, the proposed method collects a \cls token from the last layer of SST, which has been shown to hold rich localization information (Fig.\ref{fig:dino_last_head}). Discriminative proposals using an external scoring classifier are collected. The top reliable proposals are used as pseudo-labels for the localization task. Such discriminative selection allows picking potential ROIs while discarding background noise. Random selection of these proposals for training furthermore prevents overfitting to a single region and promotes exploring different parts of an object.

\section{The Proposed Discriminative Pseudo-label Sampling}
\label{sec:method}

\subsection{Notation}\label{subsec:notation}
\begin{figure*}[!t]
\begin{center}
 \includegraphics[width=1\linewidth,trim=0 0 0 570, clip]{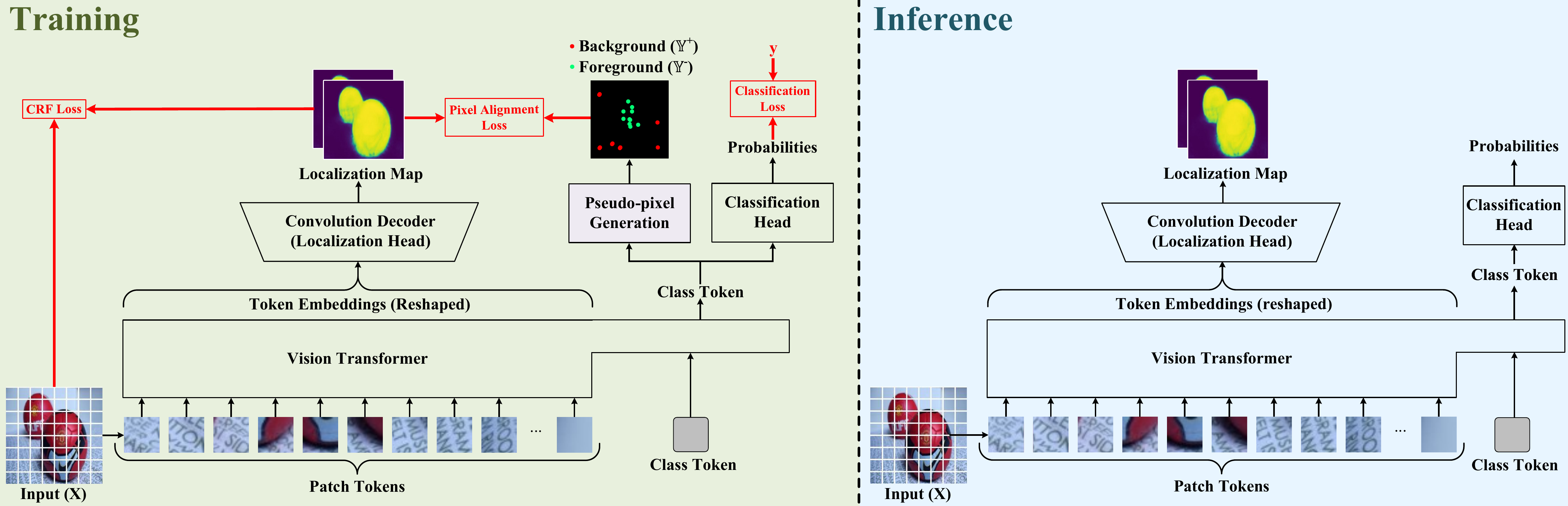}
\end{center}
\caption{\textit{DiPS}: Our proposed method for training a transformer network for WSOL tasks using a combination of localization and classification networks. The full model is composed of a vision transformer encoder, a classification head, and a localization head.
\textbf{Training}: An image class label is required to train the classification head, and our generated pixel-wise pseudo labels are needed to train the localization head. Fig.\ref{fig:pseudo_label} illustrates how this pseudo-supervision is computed.
\textbf{Inference}: In a single forward pass through our model over the input image, the classification head yields the predicted class probabilities, while the localization head localizes the object in the image.
}
\label{fig:our_method}
\end{figure*}

We denote by ${\mathbb{D} = \{(\bm{X}, y)_i\}_{i=1}^N}$ a training set, where 
${\bm{X}_i: \Omega \subset \reals^2}$ is an image, and ${\Omega}$ is a discrete image domain. The image class label is denoted as ${y_i \in \{1, \cdots, C\}}$, with $C$ being the total number of classes. Our model performs both classification and localization tasks. It is composed of three main parts (Fig.\ref{fig:our_method}):

\noindent\textbf{1) A transformer-based encoder}~\citep{dosovitskiy2020image} ${f_e}$, which takes the input image and produces token embedding ${\bm{E}}$, and $N$ \cls tokens ${\bm{E}_{cl}=\{e_0, \cdots, e_{N-1}\}}$. Equipped with its own prediction head, this model is trained in a self-supervised manner~\citep{caron2021emerging}. We keep the encoding part for our model and freeze it (no more training) for further use. Its classifier part is discarded. The parameters set of this encoder is denoted as ${\bm{\theta}_e}$.

\noindent\textbf{2) A classification head} ${f_c}$ performs classification using the \cls tokens ${\bm{E}_{cl}}$ as input. Its parameters are referred to as ${\bm{\theta}_c}$. This head produces per-class probabilities ${f_c(\bm{E}_{cl}) \in [0, 1]^C}$, where ${f_c(\bm{E}_{cl})_k = \mbox{Pr}(k | \bm{X})}$. It is trained to perform classification using standard cross-entropy: ${\min_{\bm{\theta}_c} \;  - \log(\mbox{Pr}(y | \bm{X}))}$. Once trained, its parameters, ${\bm{\theta}_c}$, are frozen.

\noindent\textbf{3) A CNN localization head} ${f_l}$, which is a decoder that performs object localization using the token embedding ${\bm{E}}$ as input. This encoder produces two full-resolution activation maps which are normalized via a softmax: ${\bm{S} = f_l(\bm{E}) \in [0, 1]^{\abs{\Omega} \times 2}}$, where ${\bm{S}^0, \bm{S}^1}$ represent the background and foreground maps, respectively. We denote by ${\bm{S}(p) \in [0, 1]^2}$ a row of matrix ${\bm{S}}$, with the index ${p \in \Omega}$ indicating a point location within ${\Omega}$. Using its own parameters ${\bm{\theta}_l}$, this module is trained to localize foreground regions over the input image ${\bm{X}}$ associated with its class ${y}$. To this end, we use our collected pseudo-labels presented in the next section.

The aforementioned three modules are part of our final model. They are used during training and inference. For our discriminative proposals sampling, we use an additional external pre-trained classifier ${f_x}$. It is used to assess the likelihood of a region of an image covering the object class associated with the image. This helps us collect reliable discriminative ROIs and discard non-discriminative background regions. It is trained over the trainset ${\mathbb{D}}$to correctly classify each sample using the image class as supervision. In this work, we use a simple the~\citep{he2016deep} network as ${f_x}$.

\subsection{Discriminative Sampling of Proposals}
\label{subs:disc_prop_samp}
In this section, we present our proposed strategy to sample discriminative proposals from a set of attention maps, \ie \cls token, ${\bm{E}_{cl}=\{e_0, \cdots, e_{N-1}\}}$, extracted from a pre-trained SST.

As illustrated in Fig.\ref{fig:dino_last_head_subfig}, self-supervised transformers produce multiple attention maps that are rich in localization information. These models tend to decompose the scene into multiple objects or parts of objects through attention maps~\citep{caron2021emerging}. While this is clearly beneficial for the localization task, attention maps and localized objects are not associated with semantic meaning, unlike CAM-based methods~\citep{oquab2015object}. Therefore, it is extremely challenging to use these maps directly for localization. To leverage these rich maps, we propose introducing a discriminative strategy to associate regions with semantic meaning, \ie the image class label. This allows for obtaining reliable discriminative ROIs that can be used as pseudo-labels for training a WSOL model.

Alg.\ref{alg:disc-prop-gen} shows a step-by-step approach to building a pool of discriminative proposals from a labeled image. Fig.\ref{fig:pseudo_label} also presents an overview of the approach. First, an external classifier, ${f_x}$ is trained using the training dataset using image class labels. This model is later used as a scoring function to measure the likelihood of a part of an image being discriminative. The process consists of iterating through all attention maps in the set ${\bm{E}_{cl}}$. Following common assumptions in CAM~\citep{zhou2016learning}, strong activations in maps are considered as potential foregrounds, while low activations are considered to be backgrounds. Therefore, the attention map is automatically thresholded~\citep{otsu1979threshold}, and a tight bounding box is computed around all connected regions.

After obtaining the set of all bounding boxes in the image from all attention maps, ${\bm{E}_{cl}}$, we proceed to score each one. We refer to this set as ${\mathbb{T}}$. To show only the contents of the box to the scoring model ${f_x}$, we perturb the image to suppress information outside the box. Particularly, blurring the outside is considered as it has proven to be more efficient in information suppression in deep models~\cite{fongPV19}. The posterior probability of the true image class is used as a score. A higher value indicates a greater likelihood that the box will contain a discriminative part related to the image class label. All boxes, ${\mathbb{T}}$, are scored and sorted. Only the boxes of the set ${\mathbb{P} = \text{top-K}(\mathbb{T})}$ are kept for further processing.
\begin{center}
\begin{minipage}{1.\linewidth}
\IncMargin{0.04in}
\removelatexerror
\begin{algorithm}[H]
    \SetKwInOut{Input}{Input}
    \SetKwInOut{Output}{Output}
    \Input{
    Input image: ${\bm{X}}$,
    \\
    Its class label: ${y}$,
    \\
    Its set of $N$ \cls tokens: ${\bm{E}_{cl}=\{e_0, \cdots, e_{N-1}\}}$,
    \\
    External pretrained classifier: ${f_x}$,
    \\
    Maximum number of proposals: ${K}$ (top-${K}$).
    }
    \vspace{0.1in}
    \Output{
    Set of top-${K}$ discriminative proposals: ${\mathbb{P}}$.
    }
    \vspace{0.03in}
    ${\mathbb{P} = \varnothing}$. \\
    \vspace{0.03in}
    Temporary proposals holder: ${\mathbb{T} = \varnothing}$. \\
    \vspace{0.1in}
    \For{\cls token ${e \in \bm{E}_{cl}}$}
    {
        Threshold the map $e$ using Otsu~\citep{otsu1979threshold} to obtain a binary map ${b}$. \\
        \vspace{0.03in}
        Find all connected region proposals in $b$, and their tight bounding boxes. Set ${\mathbb{BBOX}}$ as the set of all the bounding boxes obtained. \\
        \vspace{0.1in}
        \For{bounding box ${bx \in \mathbb{BBOX}}$} 
        {
        \textbf{Perturb the image}: Blur the content of the image \emph{outside} the bounding box ${bx}$. Leave the content of the image \emph{inside} the box. Denote the perturbed image as ${\bm{X}_{bx}}$.\\
        \vspace{0.03in}
        Compute the probability of the true class: ${f_x(\bm{X}_{bx})_y}$.\\
        \vspace{0.03in}
        Store the bounding box proposal and its probability score: ${\mathbb{T} \leftarrow \mathbb{T} \cup \{(bx, f_x(\bm{X}_{bx})_y)\}}$  
        }
    }
    \vspace{0.1in}
    Sort ${\mathbb{T}}$ using probability scores in descending order.\\
    \vspace{0.03in}
    Take the top-${K}$ most discriminative proposals: ${\mathbb{P} = \text{top-K}(\mathbb{T})}$.
    \caption{Our discriminative proposals generation (Fig.\ref{fig:pseudo_label}) from attention maps of a self-supervised transformers (SST).}
    \label{alg:disc-prop-gen}
\end{algorithm}
\DecMargin{0.04in}
\end{minipage}
\end{center}


This process of generating discriminative proposals allows the building of a rich and diverse pool of potential ROIs, which makes it more advantageous than CAM-based ROIs (Fig.\ref{fig:camfromcnn}). The latter are very local and limited to a small discriminative region, which limits their use in generating good localization pseudo-labels. In the next section, we describe how the generated bounding box proposals, ${\mathbb{P}}$, are leveraged to create pixel-wise pseudo-labels to train a WSOL model.

\subsection{Pixel-wise Pseudo-labels}
\label{subs:pixel-wise-pseudo-labels}

Using a collected set of region proposals, ${\mathbb{P}}$, we proceed to create pixel-wise pseudo-labels. In particular, we use these regions to indicate foreground and background pixels. This pseudo-supervision will later be used to train our decoder, ${f_l}$, to perform object localization following recent successful approaches over CAMs~\citep{Murtaza_2023_WACV,tcamsbelharbi2023, belharbi2023colocam,negevsbelharbi2022, belharbi2022fcam}.

The sampling of pixel supervision is tied to a bounding box. Here, we present the sampling of pixel-wise pseudo-labels for a single bounding box, assuming that ${e}$ is the attention map from the set of \cls tokens ${\bm{E}_{cl}}$ from where a bounding box was generated. Instead of considering the entire content of the box as foreground and keeping it fixed~\cite{KolesnikovL16}, a stochastic sampling of locations is considered. Such a random approach has been shown to be more efficient as it helps avoid overfitting the bounding box~\citep{belharbi2022fcam} and promotes exploring object parts. Assuming that an object is contiguous, we use the magnitude of activations inside the box to guide the sampling using a multinomial distribution. In particular, we consider the top-${n^+}$ pixels inside the box for sampling. This helps in exploring different parts of the region inside the box while focusing on the potential pixels. Among these pixels, few locations are sampled as foreground.

Sampling of background pixels is done with respect to all bounding boxes. In particular, we ensure that the sampled background pixel lands outside all the bounding boxes. Similarly to foreground pixels, we guide the sampling of the background pixels by the low magnitude activations of the map ${e}$. We sort all activations of ${e}$ and take the low-${n^-}$ pixels from which to sample. Assuming that background regions are uniformly distributed over the image, we use uniform, instead of multinomial, sampling to pick a pixel. As such, few pixels are selected to be the background.

Foreground and background pixel locations are sampled randomly. It is done at every training step for each image. This allows exploring different regions and prevents overfitting to a specific part. Additionally, it gives the decoder enough time to allow the emergence of consistent foreground regions. Since the training is done using only a few pixels at a time, the decoder learns to consistently \emph{fill in the gap} across the rest of the image by transferring the knowledge learned from the pseudo-labels to other similar regions.

The sampled pixel locations are encoded in the image domain ${\Omega^{\prime}}$. They are gathered in a partial pseudo-label mask ${\bm{Y}(p) \in \{0, 1\}^2}$ with labels ${0}$ for background and ${1}$ for foreground. Locations with unknown labels are encoded as unknown. Training the decoder is done by performing a \emph{pixel-wise alignment} between the output maps ${\bm{S}}$ and the pseudo-supervision Y. At location p, we use partial cross-entropy as follows: 

\begin{equation}
    \label{eq:pl}
    \begin{aligned}
    &\bm{H}_p(\bm{Y}, \bm{S}) =  &- (1 - \bm{Y}(p))\; \log(\bm{S}^0(p))- \bm{Y}(p) \; \log(\bm{S}^1(p)) \;, \text{for } p \in \Omega^{\prime}\;.
    \end{aligned}
\end{equation}

In practice, at each training step, and in each image, we sample several pixels from the foreground and background and ensure that they are balanced. Additionally, since the proposals are selected to be aligned with the true image class, \ie ${y}$, the selected foreground pixels follow as well. They indicate the same object annotated in the image. Therefore, the final foreground map ${\bm{S}^1}$ points similarly to the image class. This allows to build a single model for all classes.

\subsection{Training DiPS Architecture}
\label{subs:train_dips}
Our model is composed of three main parts that we train separately. The full model is illustrated in Fig.\ref{fig:our_method}.
The first module is a transformer-based encoder~\citep{dosovitskiy2020image}, ${f_e(\cdot; \bm{\theta}_e)}$. Initially, it is equipped with a classification output module. It is trained in a self-supervised fashion, similarly as in~\citep{caron2021emerging}. Once trained, the classification module is discarded, and only the encoder part is kept for our model. Its pre-trained parameters are frozen and are not trained further. In practice, we typically use pre-trained models on ImageNet dataset~\citep{RussakovskyILSVRC15} or fine-tune it on the corresponding dataset.

In addition, we have a classification module, ${f_c(\cdot; \bm{\theta}_c)}$, which relies on the encoder output features, ${\bm{E}_{cl}}$, to classify the image: ${f_c(\bm{E}_{cl})_k = \mbox{Pr}(k | \bm{X})}$. It is trained using standard cross-entropy:

\begin{equation}
    \label{eq:cl-loss}
    \min_{\bm{\theta}_c} \;  - \log(\mbox{Pr}(y | \bm{X}))\;.
\end{equation}
Once this module is trained, its weights are frozen and are no longer modified.

The last module, ${f_l(\cdot; \bm{\theta}_l)}$, is for the localization task. It is a CNN-based decoder that outputs two full-size maps for foreground and background regions. Its training loss combines two elements: pixel-wise pseudo-labels and a Conditional Random Field (CRF) loss~\citep{tang2018regularized}. To ensure that the activations of the output map ${\bm{S}}$ are well aligned with the object boundaries, a CRF loss~\citep{tang2018regularized} is employed. This loss considers both the pixels’ proximity and color similarity:
\begin{equation}
    \label{eq:crf-loss}
    \mathcal{R}(\bm{S}, \bm{X}) = \sum_{r \in \{0, 1\}} {\bm{S}^r}^{\top} \; \bm{W} \; (\bm{1} - \bm{S}^r) \;,
\end{equation}
where ${\bm{W}}$ denotes an affinity matrix in which ${\bm{W}[i, j]}$ captures the color similarity and proximity between pixels ${i, j}$ in the image ${\bm{X}}$. We employ a Gaussian kernel~\citep{koltun2011efficient} to compute ${\bm{W}}$.

The sampling of pixel-wise pseudo-labels from a single bounding box proposal was presented in a previous section (Sec.\ref{subs:pixel-wise-pseudo-labels}). However, we collected a pool of top-${K}$ bounding boxes, \ie ${\mathbb{P}}$. To avoid overfitting over a single bounding box and promote exploring the different object parts, we randomly select a bounding box from ${\mathbb{P}}$, at each training step and for each sample. Then, we proceed to sample pixel-wise pseudo-labels from the selected box. The following is the full training loss for the localization decoder for a single sample:

\begin{equation}
    \label{eq:loc-loss}
    \min_{\bm{\theta}_l} \;  \lambda_1 \; \sum_{p \in \Omega^{\prime}} \bm{H}_p(\bm{Y}, \bm{S}) + \lambda_2\; \mathcal{R}(\bm{S}, \bm{X})\;,
\end{equation}
where ${\lambda_1, \lambda_2}$ are weighing coefficients. Following~\citep{tang2018regularized}, we set $\lambda_2 = 2e^{-9}$. A validation set is used to search for ${\lambda_1}$.

\section{Results and Discussion}\label{sec:exp}

\subsection{Experimental Methodology} \label{sec:expsetting}

\noindent \textbf{Datasets:}
To validate our proposed method, we employed four challenging WSOL datasets, namely, OpenImages, ILSVRC, CUB-200-2011, and TelDrone. \textbf{(i) OpenImages} \citep{benenson2019large, choe2020evaluating} is a dataset comprising $37,319$ images divided into $100$ classes. We allocated $29,819$ images for training and $5,000$ images for testing. The remaining $2,500$ images served as the validation set. \textbf{(ii) ILSVRC} includes approximately $1.2$ million images spanning over $1,000$ classes. Following the recommendations in \citep{choe2020evaluating}, we extracted $50,000$ images for training and testing and $10,000$ images for validation. \textbf{(iii) CUB200-2011} consists of $11,788$ images distributed across $200$ categories, with $5,994$ designated for training and $5,794$ for testing \citep{WelinderEtal2010}. To perform a validation and hyperparameter search, we used an independent validation set of $1,000$ images compiled by \citep{choe2020evaluating}. \textbf{(iv) TelDrone} is a private dataset, which is owned by Ericsson Inc\footnote{\url{https://www.ericsson.com/}}. It includes $915$ high-resolution $4K$ images captured via a drone orbiting around a tower site. These images are bifurcated into two classes: one class containing images with an inspection site, and the other, those devoid of it. We divided these images into a training set of $797$ images, a validation set containing 13 images, and a test set with $105$ images.

\noindent \textbf{Evaluation measures:}
For performance evaluation, four measures and three error metrics were considered. They are commonly used in WSOL task evaluation~\citep{choe2020evaluating,gao2021ts} -- \textbf{(i)} \pxapbold denotes pixel-wise precision and recall at a particular threshold. \textbf{(ii)} \newmaxboxaccbold represents the average proportion of predicted bounding boxes with \iou values exceeding a specified threshold relative to the ground truth map. It is computed by averaging the results across three different \iou thresholds $\delta = \{30\%, 50\%, 70\%\}$. \textbf{(iii)} \toponebold localization, defined as the fraction of images, where the predicted class label correctly matches the ground truth and its corresponding \iou exceeds $\delta = 50$  \textbf{(iv)} \topfivebold localization is the fraction of images in which the ground truth class label is among the \topfive predicted class labels and the \iou is greater than $\delta =  50$. \textbf{(v) Localization part error (\lpebold)} detects instances where the localization map partially captures an object along with an intersection over the predicted bounding box (\iop) value greater than $0.5$. \textbf{(vi) Localization more error (\lmebold)} indicates that the predicted bounding box is larger than the actual bounding box, potentially including adjacent objects or background regions. This detects the instances whose intersection over the annotated-bounding box (\ioa) value is greater than $0.7$. \textbf{(vii) Multi-instance Error (\miebold)} indicates the ratio of the predicted bounding box intersecting with more than one bounding box, with an intersection-over-ground-truth-box (\iog) value exceeding $0.3$.

\noindent \textbf{Implementation details:} In all experiments, we follow the protocol proposed in \citep{choe2020evaluating}. For all datasets, we employ a batch size of $32$, resize images to $256 \times 256$, and then randomly crop to $224 \times 224$, followed by random horizontal flipping, as described in~\citep{choe2020evaluating}. We then employ the Stochastic Gradient Descent (SGD) optimizer and search the learning rate between $\{0.1,0.0001\}$. For training, we employ $50$ epochs for the CUB and TelDrone datasets and reduce the epochs to 10 for the OpenImages and ILSVRC datasets.

\noindent \textbf{Baseline Models:} To validate the performance of our proposed method, we compare our results with various state-of-the-art methods, as presented in Table \ref{tab:results_openimg_cub}. We obtain the quantitative results of CAM \citep{zhou2016learning}, ADL \citep{choe2019attention}, HaS \citep{singh2017hide}, ACoL \citep{zhang2018adversarial}, SPG \citep{zhang2018self}, and CutMix \citep{yun2019cutmix} from \citep{choe2020evaluating}. For other methods, we present the quantitative results as reported in their respective publications. Additionally, we reproduce the qualitative results of CAM \citep{zhou2016learning}, HaS \citep{singh2017hide}, ADL \citep{choe2019attention}, ACoL \citep{zhang2018adversarial}, SPG \citep{zhang2018self}, CutMix \citep{yun2019cutmix}, and TS-CAM \citep{gao2021ts} by following the protocols outlined in \citep{choe2020evaluating}. Furthermore, we compare the visual results of all datasets with the \cls tokens of the last layer of SST, which are used to collect pseudo-labels.

\subsection{Comparison with State-of-Art Methods}
\label{comparision_with_basline}

\noindent \textbf{Quantitative Comparison.} Evaluation of different localization datasets demonstrates the performance of DiPS. Table \ref{tab:results_openimg_cub} shows that DiPS surpasses the baseline model and other related methods on OpenImages when evaluated using \pxap metric. Specifically, DiPS achieves a \pxap of ${74.9\%}$, surpassing other recent methods. In particular, our method yields better results than the recent method, F-CAM~\cite{belharbi2022fcam}, with a \pxap of ${72.2\%}$, which relies on pseudo-labels from CAMs. Similarly, on the ILSVRC, CUB, and TelDrone datasets, DiPS achieves competitive performance in terms of \newmaxboxacc, \topone and \topfive localization accuracy metrics, as shown in Table \ref{tab:results_openimg_cub}, \ref{tab:results_ilsvrc} and \ref{tab:teldrone}. Additionally, we also present the localization accuracy at different thresholds $\delta$ on the ILSVRC dataset to show that our approach outperforms the other methods with a high \iou value. This indicates that our method produces a more accurate localization, as demonstrated in the following qualitative evaluation. Qualitative results also show that our activation maps have sharp boundaries.

\begin{figure*}[!ht]
\begin{center}
 \includegraphics[width=0.8\linewidth,trim=7 7 7 7, clip]{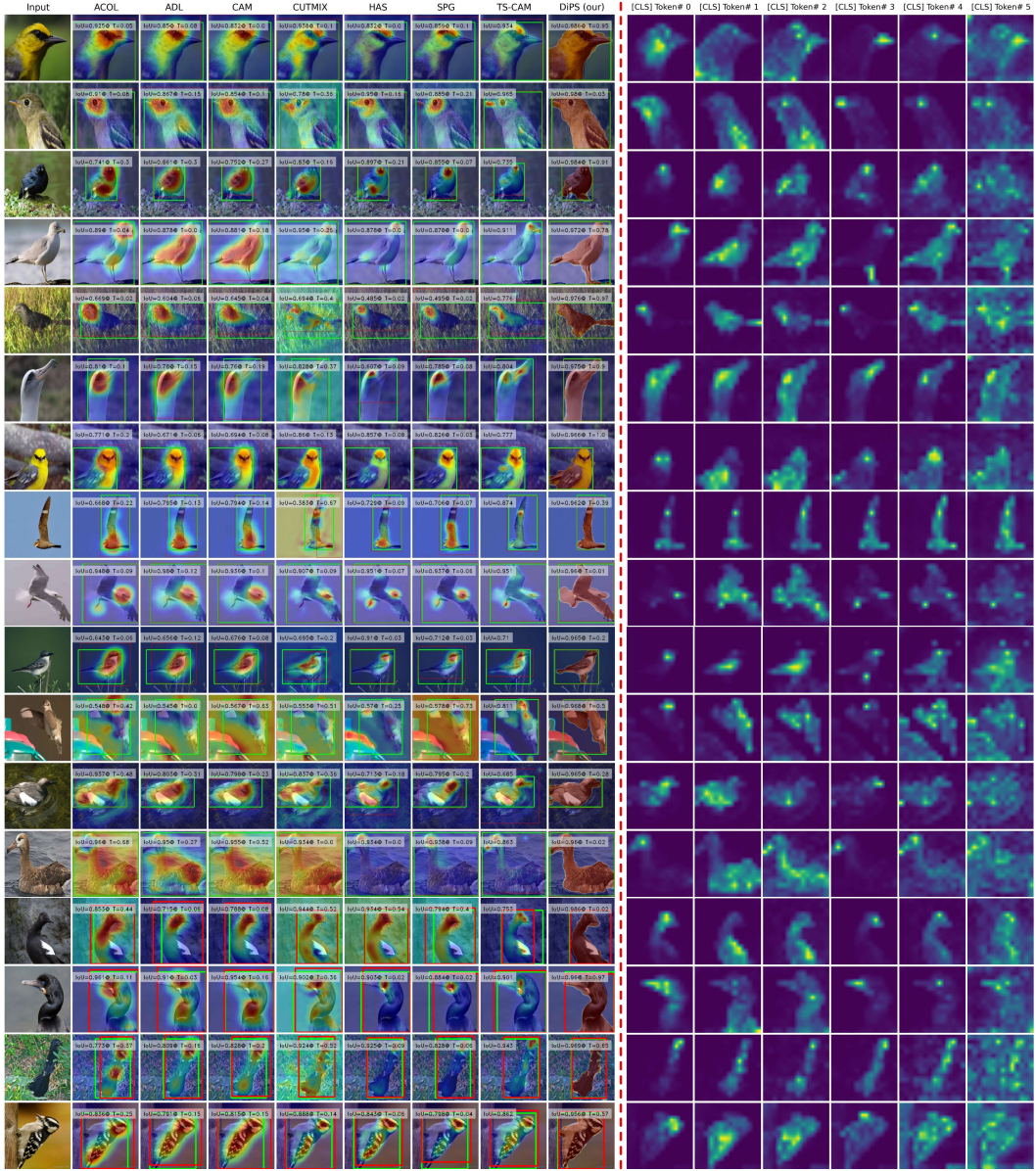}
\end{center}
\caption{Samples from the test set on the CUB-200-2011 dataset generated by our proposed and state-of-the-art methods. In different examples, the results of baseline methods highlight different parts of the object, but the entire object is encompassed within the bounding box due to an extensive search of thresholds. In contrast, the map generated by our method covers the full object with relatively consistent activation values across the object. Additionally, our model effectively mitigates noise in the generated map, which is present in the \cls tokens used to produce pseudo-labels.  Here, {\color{green}green} denotes the ground truth bounding box, while {\color{red}red} corresponds to the predicted bounding box.}\label{fig:cub_results_v1}
\end{figure*}

\begin{figure*}[!b]
\begin{center}
 \includegraphics[width=0.7\linewidth,trim=6 6 6 6, clip]{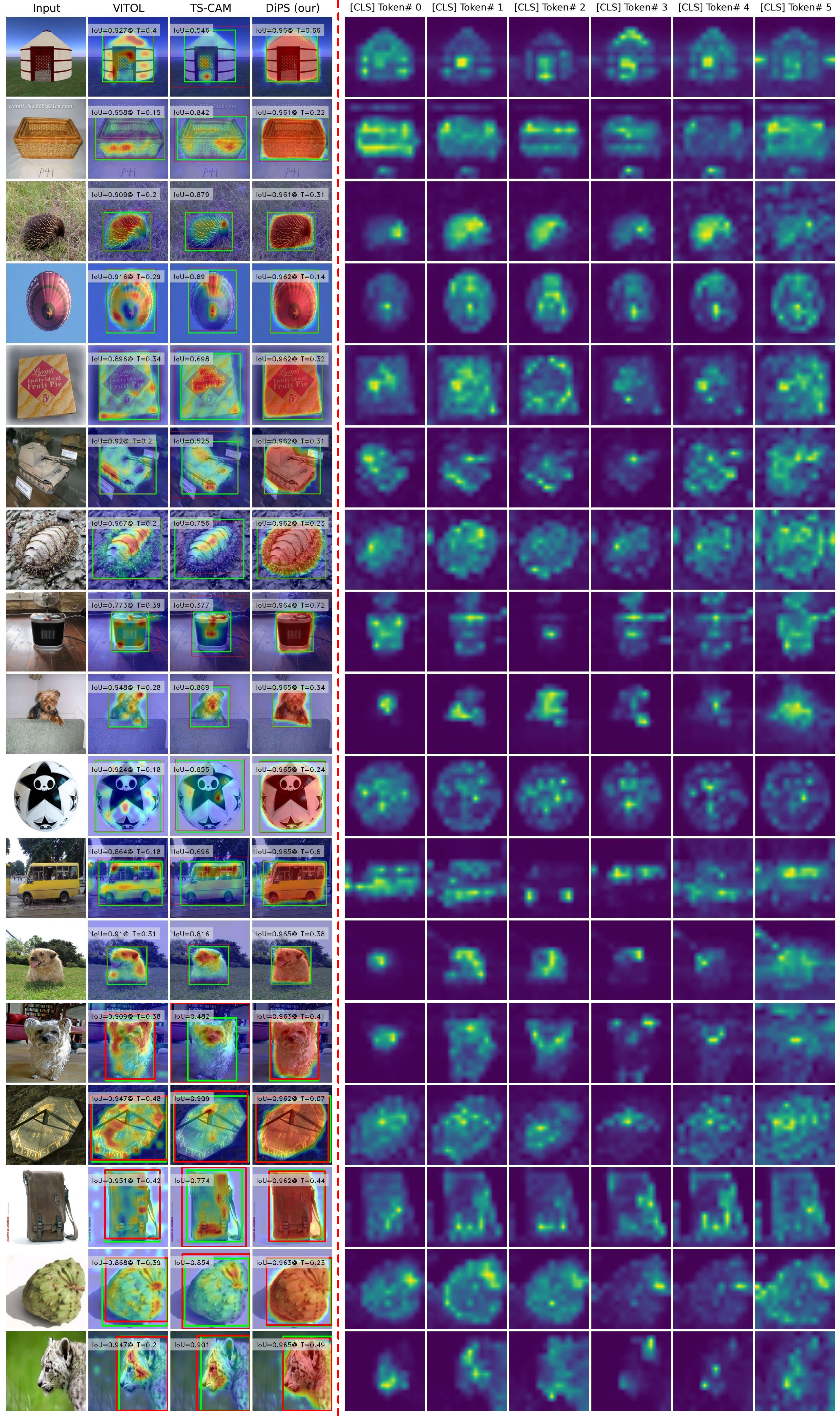}
\end{center}
\caption{Visual results for proposed and state-of-the-art methods on the ILSVRC dataset.  Here, {\color{green}green} denotes the ground truth bounding box, while {\color{red}red} corresponds to the predicted bounding box.}\label{fig:ilsvrc_res}
\end{figure*}

\begin{figure*}[!t]
\begin{center}
 \includegraphics[width=0.64\linewidth,trim=6 229 6 6, clip]{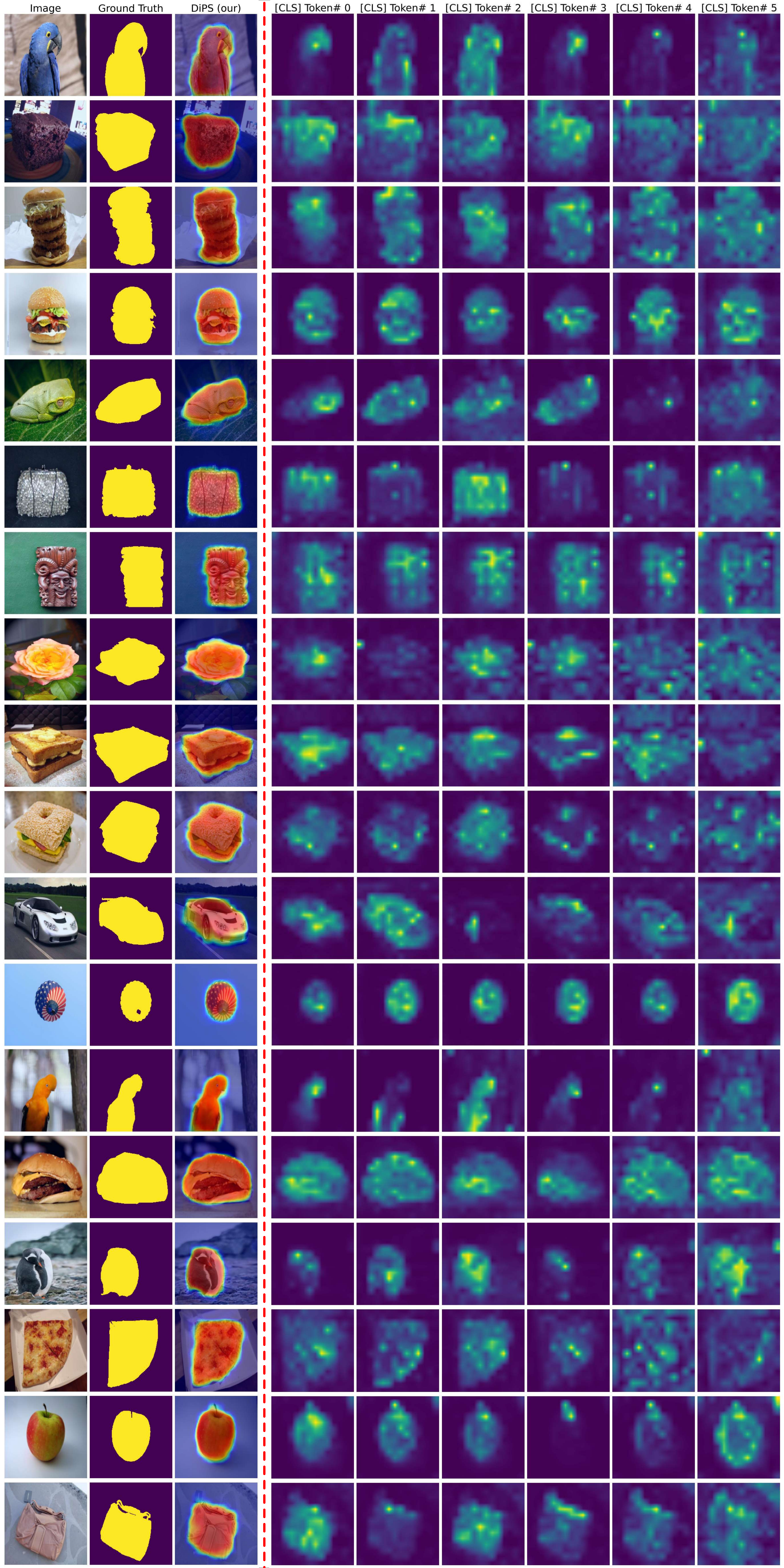}
\end{center}
\caption{Visualization of results on the OpenImages dataset. }\label{fig:openimg_res}
\end{figure*}

\begin{figure}[!ht]
\begin{center}
 \includegraphics[width=0.94\linewidth,trim=6 6 6 6, clip]{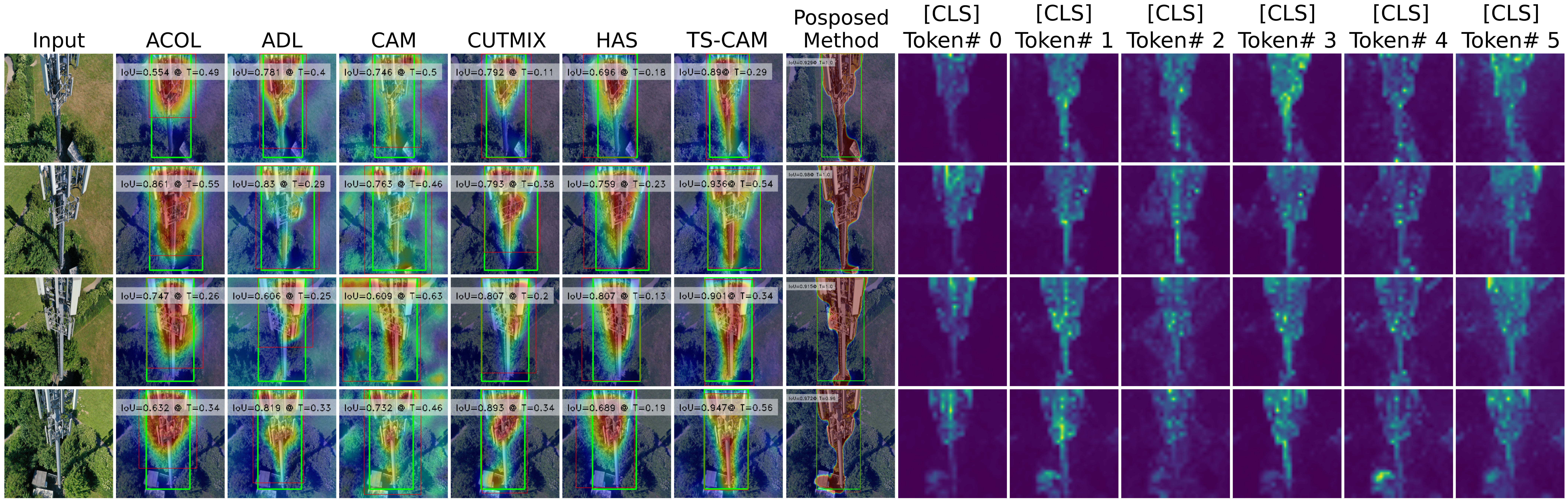}
\end{center}
\caption{Visualization of results on the TelDrone dataset. We added a few examples for this dataset, given the proprietary restrictions specified by Ericsson Incorporation. Here, {\color{green}green} denotes the ground truth bounding box, while {\color{red}red} corresponds to the predicted bounding box.}\label{fig:ericsson_res}
\end{figure}

\noindent \textbf{Qualitative Comparison.}
Visual results of DiPS and related baselines are presented in Fig.\ref{fig:cub_results_v1}-\ref{fig:ericsson_res}, indicating that our method yields good localization. Baseline methods focus on common areas of discriminative regions that are shared among different objects of the same class, as they highlight these areas. The blobby nature of these maps necessitates the selection of optimal threshold values to accurately identify a particular object. However, this thresholding mechanism renders localization maps unreliable and may include regions with concealed activations for an object of interest. In contrast to the baseline models, DiPS is able to highlight foreground objects, maintaining a uniform activation across different object parts. For instance, on the CUB-200-2011 dataset, our method produces robust activation maps with sharper boundaries, outperforming current state-of-the-art techniques. Additionally, our method maintains a consistent localization performance across different datasets, including ILSVRC, OpenImages, and TelDrone. For instance, DiPS is able to distinguish the foreground object even in scenarios where the object’s texture closely resembles the background regions(Fig.\ref{fig:openimg_res}). Additionally, we also compared the output of our method with the \cls tokens used to collect pseudo-labels (Fig.\ref{fig:cub_results_v1}-\ref{fig:openimg_res}). These visualizations indicate that our method is able to identify objects efficiently while mitigating the noise that is present in the \cls token of the last attention blocks.

\input{helperfiles/table1}
\input{helperfiles/table2}
\input{helperfiles/table6}

All in all, state-of-the-art methods generate localization maps for objects of interest with varying intensities. They are also able to properly draw a bounding box around the object due to an extensive threshold search, which can include low-scoring areas in the localization map. In contrast, DiPS produces localization maps that identify a particular object with sharper boundaries, thus eliminating the need for a precise threshold value.

\noindent \textbf{Comparison with Self-Supervised Methods.} In addition to the above results, we also report the \newmaxboxacc performance of different self-supervised vision transformers, including SST \citep{caron2021emerging}, used to collect pseudo-labels to train our model (Table \ref{tab:clip-es}).To extract localization maps from these models, we gather different attention maps and overlay them to the original image to produce different perturbed images. One of the perturbed images is selected based on the classifier’s score and its corresponding attention map to compute the localization performance. Moreover, we also report the \newmaxboxacc performance of maps extracted from the Contrastive Language-Image Pretraining (CLIP-ES) model~\citep{lin2023}. CLIP-ES is a prompt-based model trained on paired text and image datasets. The CLIP-ES model requires a class label alongside the input image to localize the ROI for the specified class. This distinguishes CLIP from traditional WSOL models, which only rely on the image for inference CAMs. The design of CLIP inherently confers upon it a significant advantage, as it is explicitly aware of the target class during its computations. In our experiments, we input a predicted class label from a pretrained classifier, along with the image, to the CLIP model. 

\begin{figure*}[!t]
\begin{center}
 \includegraphics[width=0.94\linewidth,trim=16 16 16 16, clip]{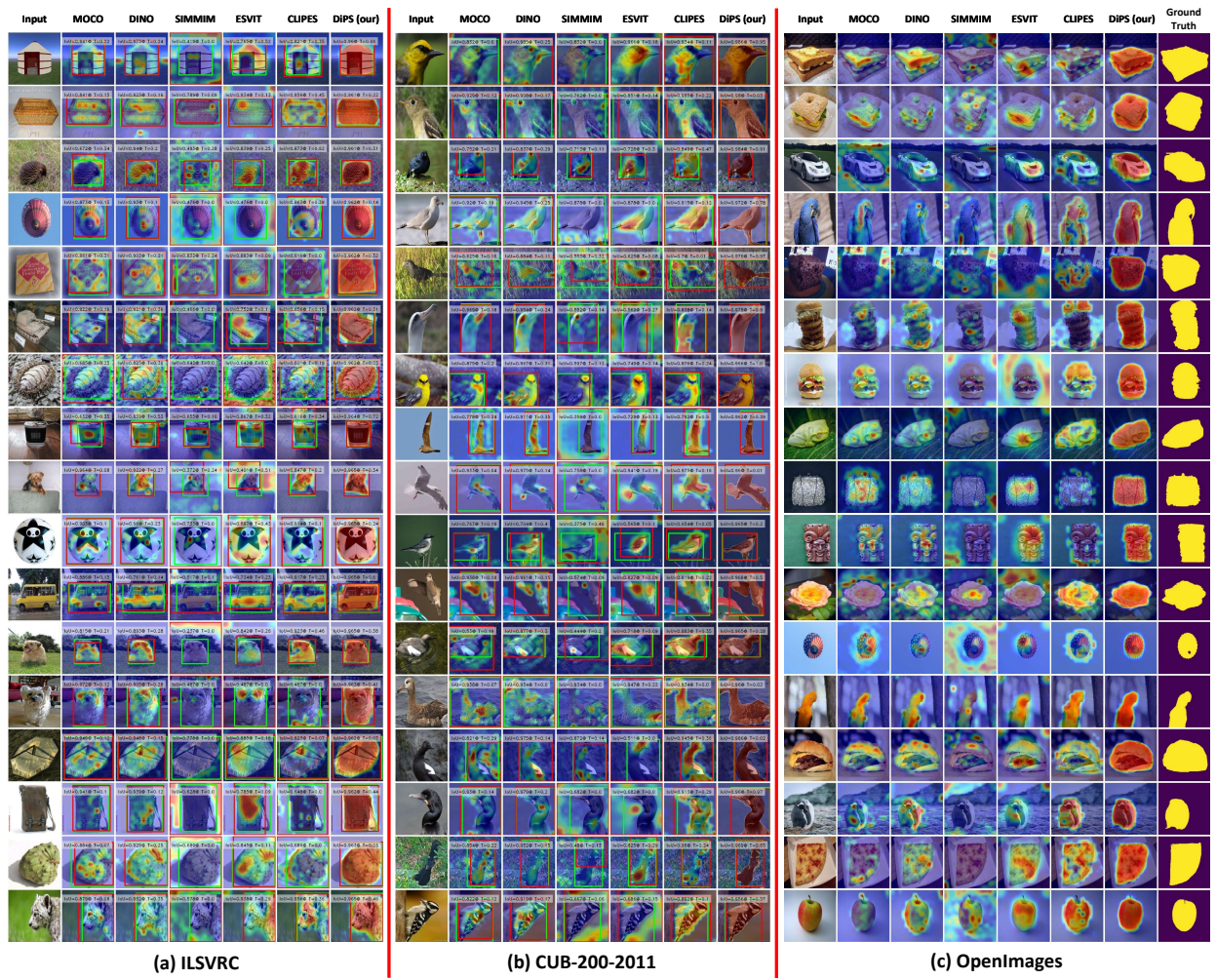}
\end{center}
\caption{Visual results of our methods compared to the self-supervised method. Here, {\color{green}green} denotes the ground truth bounding box, while {\color{red}red} corresponds to the predicted bounding box.}\label{fig:ssl_result}
\end{figure*}

Empirical results in Table \ref{tab:clip-es} indicate that our model is capable of achieving high performance compared to the self-supervised methods, using only class-level labels in terms of \newmaxboxacc. Visual results show that, in contrast to our method, self-supervised methods hotspot a different region, which requires an optimal threshold to draw an optimal bounding box, as shown in Fig.\ref{fig:ssl_result}. This analysis not only enriches the comparative analytical landscape but also fortifies the benefits of DiPS as a robust and generalizable methodology for object localization.

\input{helperfiles/table4_revised}

\noindent\textbf{Complexity Analysis.} DiPS achieves competitive performance in comparison to the baseline model introduced in our preliminary study \citep{Murtaza_2023_WACV}. The model presented in this paper requires a single forward pass to generate both classification scores and localization maps. Moreover, the forward pass of our model heads (localization and classification) requires merely 2.396G Multiply-Accumulate (MACs) operations with 1.616M parameters, as opposed to our previous model~\citep{Murtaza_2023_WACV}, which called for 31.895G MACs with 58.078M parameters for inference.

\begin{figure}[!ht]
\begin{center}
 \includegraphics[width=1\linewidth,trim=3 3 3 3, clip]{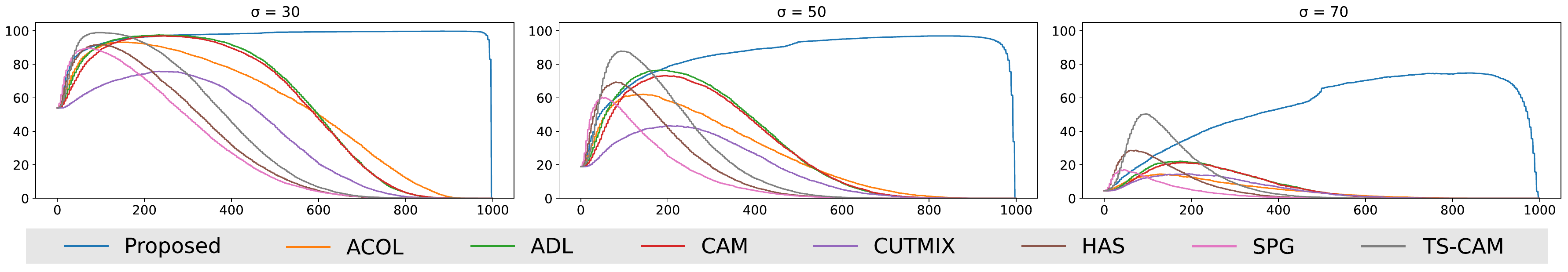}
\end{center}
\caption{The \maxboxacc performance of DiPS and state-of-the-art methods is at different threshold values ($\delta=\{30,40,70\}$) calculated on 
a test set from the CUB-200-2011 dataset.
}\label{fig:acc_over_ths_v1}
\end{figure}

\noindent\textbf{Analysis of Distribution Shift.}
In this section, we analyze the impact of varying thresholds on the localization performance, along with the distribution of activation values corresponding to the object of interest. The shift in \maxboxacc at different threshold values for our model selected during the hyperparameter search through \maxboxacc is presented in Fig.\ref{fig:acc_over_ths_v1}. For the baseline methods, we observe that the \maxboxacc rapidly declines to zero as the threshold increases. This makes it challenging to search for the optimal threshold values for each image during inference. In contrast, the output generated by our method exhibits a lower susceptibility to threshold variations.

\subsection{Error Analysis and Ablation Study}
For a fair evaluation of our presented approach, different error metrics adopted from \citep{gao2021ts} are employed (Section \ref{sec:expsetting}). Using these metrics, we analyze the performance of our model on the CUB-200-2011 and ILSVRC datasets (Table \ref{tab:abla_study_cub_ilsvrc}). These results show that the localization maps produced by our method are able to accurately localize a particular object, avoiding any overestimation or underestimation of the object of interest. In contrast to the related methods, \mie on the ILSVRC dataset demonstrates the ability of our method to localize a specific object rather than multiple objects. This analysis also reveals that the maps generated by our method are highly robust and exhibit considerably fewer errors as compared to the baseline methods.

Furthermore, the performance of our model, selected during the hyperparameter search without a CRF loss, is presented in Table \ref{tab:abla_study_cub_ilsvrc}. This indicates that CRF loss significantly contributes to the model’s performance. Without the CRF loss, our pixel-alignment loss exhibits a slightly lower performance as compared to the results reported in the previous section. Thus, the inclusion of the CRF loss notably enhances our model’s performance.

\input{helperfiles/table5}
\input{helperfiles/table3}

\section{Conclusion}\label{sec:conc}

In this paper, we proposed a novel transformer-based method for the WSOL task. In particular, we designed a discriminative approach to sample reliable proposals from the attention maps of a self-supervised transformer. Such maps have proven to be rich with localization information but lack semantic meaning. Using a pre-trained classifier, we score region proposals and measure their likelihood of containing the true image class. Only top-scoring proposals are retained. Our sampling strategy allows us to build a diverse and rich pool of region proposals to train a WSOL model.
Additionally, we design a transformer-based model for WSOL that aims to achieve the best performance for both classification and localization tasks simultaneously.
Experimental results demonstrate that our method is capable of producing reliable localization maps, outperforming the \cls tokens used to generate pseudo-labels. Our model generates robust localization maps that exhibit less sensitivity to threshold values. Results of our method over four challenging datasets show its benefits compared to state-of-the-art methods. Furthermore, our model can produce localization maps with consistent intensities across all object parts, unlike its counterpart methods.

\clearpage

\noindent \textbf{Acknowledgements:} This research was supported by the \textit{Mathematics of Information Technology and Complex Systems} and the \textit{Natural Sciences and Engineering Research Council of Canada}. We also acknowledge \textit{Digital Research Alliance of Canada} for their provision of computing resources.

{\small
\bibliographystyle{unsrtnat}
\bibliography{biblio}
}
\end{document}

%% file: helperfiles/table1.tex
\begin{table}[!ht]
\centering
\resizebox{0.87\linewidth}{!}{%
                    \begin{tabular}{l|ccccccccc}
\toprule
&  & & && \textbf{OpenImages} && \multicolumn{3}{c}{\textbf{CUB}}\\
 \cline{6-6}\cline{8-10}   
\textbf{Method} & & \multicolumn{2}{c}{\textbf{\backbonebold}} &&  \textbf{\pxapbold} && \textbf{\newmaxboxaccbold} & \toponebold \locbold & \topfivebold \locbold \\
\hline\hline
CAM~\citep{zhou2016learning} {\small \emph{(cvpr,2016)}} && \multicolumn{2}{c}{ResNet50} & & 63.2 && 63.7 & 56.1 & -- \\
HaS~\citep{singh2017hide} {\small \emph{(iccv,2017)}} && \multicolumn{2}{c}{ResNet50}  & & 58.1 && 64.7 & 60.7 & -- \\
ACoL~\citep{zhang2018adversarial} {\small \emph{(cvpr,2018)}} && \multicolumn{2}{c}{ResNet50}  & & 57.3 && 66.5 & 57.8 & -- \\
SPG~\citep{zhang2018self} {\small \emph{(eccv,2018)}} && \multicolumn{2}{c}{ResNet50}  & & 62.3 && 60.4 & 51.5 & -- \\
ADL~\citep{choe2019attention} {\small \emph{(cvpr,2019)}} && \multicolumn{2}{c}{ResNet50}  & & 58.7 && 66.3 & 41.1 & -- \\
CutMix~\citep{yun2019cutmix} {\small \emph{(eccv,2019)}} && \multicolumn{2}{c}{ResNet50}  & & 62.5  && 62.8 & 54.5 & -- \\
PAS~\citep{bae2020rethinking} {\small \emph{(eccv,2020)}} && \multicolumn{2}{c}{GoogleNet}  & & 63.3 && -- & -- & -- \\
PAS~\citep{bae2020rethinking} {\small \emph{(eccv,2020)}} && \multicolumn{2}{c}{ResNet50}  & & 60.9 && -- & -- & -- \\
ICL~\citep{ki2020insample} {\small \emph{(accv,2020)}} && \multicolumn{2}{c}{ResNet50}  & & -- && 63.1 & 56.1 & --  \\
CAM-IVR~\citep{kim2021normalization} {\small \emph{(iccv,2021)}} && \multicolumn{2}{c}{InceptionNet}  & & 63.6 && 66.9 & -- & -- \\
CAM-IVR~\citep{kim2021normalization} {\small \emph{(iccv,2021)}} && \multicolumn{2}{c}{ResNet50}  & & 58.9 && 60.9 & -- & -- \\
TS-CAM~\citep{gao2021ts} {\small \emph{(iccv,2021)}} && \multicolumn{2}{c}{DeiT-S}  & & -- && 76.7 & 71.3 & 83.8 \\
ViTOL-GAR~\citep{gupta2022vitol} {\small \emph{(cvpr,2022)}} && \multicolumn{2}{c}{DeiT-S}  & & -- && 72.4 & -- & -- \\
ViTOL-LRP~\citep{gupta2022vitol} {\small \emph{(cvpr,2022)}} && \multicolumn{2}{c}{DeiT-S}  & & -- && 73.1 & -- & -- \\
PDM~\citep{meng2022diverse} {\small \emph{(tip,2022)}} && \multicolumn{2}{c}{ResNet50}  && -- & & 72.4 & -- & -- \\
C$^2$AM~\citep{xie2022c2am} {\small \emph{(cvpr,2022)}} && \multicolumn{2}{c}{ResNet50} & & -- && 83.8 & 76.6 & 89.15 \\
SCM~\citep{bai2022weakly} {\small \emph{(eccv,2022)}} && \multicolumn{2}{c}{DeiT-S}  & & -- && 89.9 & -- & -- \\
TRT~\citep{su2022re} {\small \emph{(corr,2022)}} && \multicolumn{2}{c}{DeiT-B}  & & -- && 82.0 & 76.5 & 88.0 \\
BGC~\citep{kim2022bridging} {\small \emph{(cvpr,2022)}} && \multicolumn{2}{c}{ResNet50}  & & -- && 75.9 & 73.2 & 86.7 \\
BGC~\citep{kim2022bridging} {\small \emph{(cvpr,2022)}} && \multicolumn{2}{c}{VGG16}  & & -- && 80.1 & 70.8 & 88.1 \\
BR-CAM~\citep{zhu2022bagging} {\small \emph{(eccv,2022)}} && \multicolumn{2}{c}{ResNet50}  & & 67.6 && -- & -- & -- \\
CREAM~\citep{xu2022cream} {\small \emph{(cvpr,2022)}} && \multicolumn{2}{c}{ResNet50}  & & 64.7 && 73.5 & 76.0 & -- \\
F-CAM+XGradCAM~\citep{belharbi2022fcam} {\small \emph{(wacv,2022)}} && \multicolumn{2}{c}{VGG16}  & & 69.0&& 80.1 & 22.0 & 49.6 \\
F-CAM+LayerCAM~\citep{belharbi2022fcam} {\small \emph{(wacv,2022)}} && \multicolumn{2}{c}{ResNet50}  & & 72.2 && 82.7 & 47.7 & 76.1 \\
\hline
\textbf{DiPS (ours)} && \multicolumn{2}{c}{DeiT-S} & & \textbf{74.9} && \textbf{91.5} & \textbf{79.2} & \textbf{92.2} \\
\bottomrule
\end{tabular}
                }
                \captionof{table}{\newmaxboxacc and \pxap performance of proposed and state-of-the-art methods on the CUB and OpenImages datasets.}
                \label{tab:results_openimg_cub}
\end{table}

%% file: helperfiles/table2.tex
\begin{table}[!ht]
\centering
\resizebox{0.77\linewidth}{!}{%
\begin{tabular}{l|cccccccgccc}
 \toprule    
& & &&& \multicolumn{4}{c}{\textbf{\newmaxboxacc}} && \multicolumn{2}{c}{\locbold}\\
\cline{6-9}\cline{11-12}
\textbf{Method} && \multicolumn{2}{c}{\textbf{\backbonebold}}  && \textbf{\boldcourier{$\delta$} = 0.3} & \textbf{\boldcourier{$\delta$} = 0.5} & \textbf{\boldcourier{$\delta$} = 0.7} & \textbf{Mean} && \toponebold & \topfivebold
\\
\hline\hline
CAM~\citep{zhou2016learning} {\small \emph{(cvpr,2016)}} && \multicolumn{2}{c}{ResNet50} & & 83.7 & 65.7 & 41.6 & 63.7 && 51.8 & -- \\
HaS~\citep{singh2017hide} {\small \emph{(iccv,2017)}} && \multicolumn{2}{c}{ResNet50} & & 83.7 & 65.2 & 41.3 & 63.4 && 49.9 & -- \\
SPG~\citep{zhang2018self} {\small \emph{(eccv,2018)}} && \multicolumn{2}{c}{ResNet50} & & 83.9 & 65.4 & 40.6 & 63.7 && 47.4 & -- \\
ADL~\citep{choe2019attention} {\small \emph{(cvpr,2019)}} && \multicolumn{2}{c}{ResNet50} & & 83.6 & 65.6 & 41.8 & 63.7 && 48.5 & -- \\
CutMix~\citep{yun2019cutmix} {\small \emph{(eccv,2019)}} && \multicolumn{2}{c}{ResNet50} & & 83.6 & 65.6 & 41.8 & 63.7 && 51.5 & -- \\
ICL~\citep{ki2020insample} {\small \emph{(accv,2020)}} && \multicolumn{2}{c}{ResNet50} & & 84.3 & 67.6 & 43.6 & 65.2 && 48.4 & -- \\
BGC~\citep{kim2022bridging} {\small \emph{(cvpr,2022)}} && \multicolumn{2}{c}{ResNet50} & & \textbf{86.7} & \textbf{71.1} & 48.3 & \textbf{68.7} && 53.8 & 65.7 \\
\hline
\textbf{DiPS (ours)} && \multicolumn{2}{c}{DeiT-S} & & 83.2 & 69.7 & \textbf{51.7} & 68.2 && \textbf{56.4} & \textbf{66.7} \\
\bottomrule
\end{tabular}
}
\caption{\newmaxboxacc performance of proposed and state-of-art methods on the ILSVRC dataset, along with IoU at different threshold values, denoted by $\delta$. We adopted some of the results for baseline methods from \citep{kim2022bridging}.}\label{tab:results_ilsvrc}
\end{table}

%% file: helperfiles/table6.tex
\begin{table}[!ht]
\centering
\resizebox{0.42\linewidth}{!}{%
\centering
\begin{tabular}{l|cccccc}
\toprule
                 &  & \multicolumn{3}{c}{\textbf{\newmaxboxacc}} \\
\textbf{Methods} &  & \textbf{VGG16} && \textbf{ResNet50}\\
\hline\hline
CAM~\cite{zhou2016learning} {\small \emph{(cvpr,2016)}}      && 55.9  &&  45.2 \\
HaS~\cite{singh2017hide} {\small \emph{(iccv,2017)}}         && 60.3  &&  49.1   \\
ACoL~\cite{zhang2018adversarial} {\small \emph{(cvpr,2018)}} && 43.4  &&  59.1 \\
SPG~\cite{zhang2018self} {\small \emph{(eccv,2018)}}         && 62.8  &&  67.3  \\
ADL~\cite{choe2019attention} {\small \emph{(cvpr,2019)}}     && 66.0  &&  63.5  \\
CutMix~\cite{yun2019cutmix} {\small \emph{(eccv,2019)}}      && 57.2  &&  50.3  \\
TS-CAM~\cite{gao2021ts} {\small \emph{(iccv,2021)}}          && \multicolumn{3}{c}{DeiT-S: 72.2}\\
\cline{1-2} \cline{3-5}
\textbf{DiPS (ours)} & & \multicolumn{3}{c}{DeiT-S: \textbf{92.4}}  \\
\bottomrule
\end{tabular}
}
\caption{Quantitative comparison of DiPS and existing state-of-the-art methods on the TelDrone Dataset using \newmaxboxacc metric. Experimental results on this proprietary dataset indicate robust performance across various datasets.
}
\label{tab:teldrone}
\end{table}

%% file: helperfiles/table4_revised.tex
\definecolor{lightgray}{rgb}{0.77,0.77,0.77}
\definecolor{gray}{rgb}{0.66,0.66,0.66}
\begin{table}[!ht]
\centering
\resizebox{.6\linewidth}{!}{%
\centering
\begin{tabular}{l|ccccccc}
\hline
        &&   \multicolumn{3}{c}{\textbf{\newmaxboxacc}} & \pxapbold \\
\textbf{Method}  &&  \textbf{CUB-200-2011}  & \textbf{ILSVRC} && \textbf{OpenImages} \\
\hline \hline 
DINO (DeiT-S) \cite{caron2021emerging}         &&  65.1   & 59.9  && 52.5  \\
ESVIT (SWIN-ViT) \cite{li2021efficient}          &&  51.5   & 54.5  && 39.1 \\
MoCoV3 (ViT-B) \cite{chen2021empirical}       &&  47.3   & 56.1 && 44.4 \\
SimMIM (ViT-B) \cite{xie2022simmim}           &&  31.2   & 44.6 && 22.1 \\
CLIP-ES \cite{lin2023}                &&  79.9   & 61.7  && 54.2  \\
\hline
\textbf{DiPS (DeiT-S) - ours}            &&  \textbf{91.5}   & \textbf{68.2} 
&& \textbf{74.3} \\
\hline
\end{tabular}
}
\caption{A comparative analysis of our approach against self-supervised and prompt-based models; In self-supervised transformers, \cls tokens are utilized to compute localization performance. For this purpose, we harvest \cls tokens from the last attention block of the transformer. These \cls tokens are then binarized and used to perturb the original image by blurring the background regions. Subsequently, this perturb is processed by a classifier, and a map with the maximum classifier score is selected for computation localization scores. This selection strategy closely adheres to the pseudo-label generation process employed in our method. For CLIP-ES, class labels are obtained from pre-trained models that are then passed through the CLIP model along with the input image to obtain attention map \cite{lin2023}. Moreover, our method is able to surpass self-supervised vision transformers and prompt-based models.
}
\label{tab:clip-es}
\end{table}

%% file: helperfiles/table5.tex
\begin{table}[h]
\centering
\resizebox{0.6\linewidth}{!}{
\begin{tabular}{l|ccccccc}
\toprule
    \textbf{Methods} && \multicolumn{2}{c}{\textbf{CUB}} &&
    \multicolumn{3}{c}{\textbf{ILSVRC}}\\
     && \lpebold$\downarrow$ & \lmebold$\downarrow$ && \miebold$\downarrow$ & \lpebold$\downarrow$ & \lmebold$\downarrow$\\
     \hline \hline
    VGG16 (CAM)  &&21.91 &10.53 && 10.65 & 3.85 & 9.58\\
    InceptionV3 (CAM) &&23.09 &5.52 && 10.36 & 3.22 & 9.49  \\
     TS-CAM \citep{gao2021ts}  && 6.30 &2.85  && 9.13 & 3.78 & 7.65   \\
     \hline
     DiPS (our) && .002  & .001 && 0.03 & 0.01 & 0.04 \\ 
	\bottomrule
\end{tabular}
}

\captionof{table}{Error analysis of our method. Results of baseline methods (VGG16, InceptionV3, TS-CAM) are borrowed from \citep{gao2021ts}.}\label{tab:error_analysis_cub_ilsvrc}
\end{table}

%% file: helperfiles/table3.tex
\begin{table}[h]
                \centering
\resizebox{0.68\linewidth}{!}{%
\begin{tabular}{l|cccccc}
\toprule
     \textbf{Losses} &&  \textbf{CUB} &&  \textbf{ILSVRC} &&  \textbf{OpenImages}\\
     && (\newmaxboxaccbold) && (\newmaxboxaccbold) && (\newmaxboxaccbold)\\
     \hline \hline
    Pixel pseudo-labels && 89.2 && 67.2 && 70.9 \\
    Pixel pseudo-labels + CRF && 91.5 && 68.2 && 74.9 \\
 \bottomrule
\end{tabular}
}
\captionof{table}{Ablation study of our model for loss functions of localization head.}
\label{tab:abla_study_cub_ilsvrc}
\end{table}